\documentclass{article}
\usepackage{arxiv}

\usepackage{amsmath, amsfonts, amsthm}
\usepackage{nicefrac}   
\usepackage{graphicx}
\usepackage{booktabs}
\usepackage{multirow}
\usepackage{threeparttable}
\usepackage{tikz}
\usepackage{subcaption}
\usepackage{hyperref}
\usepackage{ifthen}
\usepackage{pifont}
\usepackage{multicol}
\usepackage{tikz}
\usepackage{pgfplots}
\pgfplotsset{compat=1.18}

\usetikzlibrary{arrows.meta,positioning,fit,shapes.geometric,calc, decorations.pathreplacing}

\usepackage[ruled,vlined,linesnumbered]{algorithm2e}

\newcommand{\ie}{\textit{i.e.}}
\newcommand{\eg}{\textit{e.g.}}

\newtheorem{theorem}{Theorem}[section]
\newtheorem{corollary}[theorem]{Corollary}
\newtheorem{lemma}[theorem]{Lemma}
\newtheorem{definition}{Definition}[section]

\newcommand{\stepcirc}[1]{%
	\tikz[baseline=(char.base)]{
		\node[draw, 
		circle,
		fill=black!15,       
		inner sep=3pt, 
		font=\sffamily \mdseries] (char) {#1};
	}%
}

\newcommand{\smallstepcirc}[1]{%
	\tikz[baseline=(char.base), remember picture]{
		\node[draw, 
		circle,
		fill=black!30,       
		inner sep=1pt, 
		font=\sffamily] (char) {#1};
	}%
}

\newcommand{\system}{\textsc{GRANITE}\xspace}

\makeatletter
\@addtoreset{footnote}{page}
\makeatother

\title{\system~: a Byzantine-Resilient Dynamic Gossip Learning Framework}
\date{}

\author{ \href{0000-0003-3433-1337}{\includegraphics[scale=0.06]{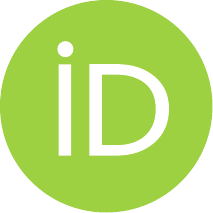}\hspace{1mm}Yacine Belal}\\
	CEA, List, Université Paris-Saclay\\ Palaiseau, France\\
	\texttt{yacine.belal@cea.fr} \\
	\And
	\href{https://orcid.org/0000-0001-6473-7173}{\includegraphics[scale=0.06]{orcid.pdf}\hspace{1mm}Mohamed Maouche} \\
	INRIA, INSA Lyon, CITI, UR3720\\ Villeurbanne, France \\
	\texttt{mohamed.maouche@inria.fr} \\
	\And 
	\href{https://orcid.org/0000-0003-2821-7714}{\includegraphics[scale=0.06]{orcid.pdf}\hspace{1mm}Sonia Ben Mokhtar}\\
	LIRIS, INSA Lyon, CNRS\\
	Lyon, France\\
	\texttt{sonia.ben-mokhtar@cnrs.fr} \\}

\begin{document}

\maketitle

\begin{abstract}
 Gossip Learning (GL) is a decentralized learning paradigm where users iteratively exchange and aggregate models with a small set of neighboring peers. Recent approaches rely on dynamic communication graphs built using Random Peer Sampling (RPS) protocols which have been proven to accelerate convergence. However, we show that these approaches are vulnerable to a dual attack: Byzantine nodes can poison models and manipulate peer sampling to amplify their influence. We address this combination of threats with \system, a framework for robust learning over sparse, dynamic graphs in the presence of Byzantine nodes. \system accumulates knowledge about encountered node identifiers over time and dynamically adjusts local aggregation thresholds based on estimated Byzantine density in the neighbourhood of each node. We demonstrate that under \system, the Byzantine presence in local neighborhoods exhibits an exponential decay. We further derive the robustness conditions of the graphs generated by \system. Empirically, our results indicate that \system converges within 5\% of non-Byzantine accuracy under 30\% Byzantines nodes, offers faster convergence and operates on graphs with up to 9x lower communication cost.
\end{abstract}


\newcommand{\argmin}[1]{\mathop{\mathrm{arg\,min}}\limits_{#1}}

\newcommand{\argmax}[1]{\mathop{\mathrm{arg\,max}}\limits_{#1}}

\newcommand{\mixing}{\textbf{W}}
\newcommand{\adjacency}{\textbf{A}}
\newcommand{\degree}{\textbf{D}}
\newcommand{\identity}{\textbf{I}}
\newcommand{\matrice}{\textbf{X}}
\newcommand{\matricep}[1]{X^{#1}}

\newcommand{\wakeup}{\mathcal{U}}
\newcommand{\users}{\mathcal{N}}
\newcommand{\nbusers}{n}
\newcommand{\round}{t}
\newcommand{\roundn}{T_0}
\newcommand{\dataset}{D}
\newcommand{\datasetu}[1]{\dataset_{#1}}
\newcommand{\distribution}{\mathcal{D}}
\newcommand{\distributionV}[1]{\distribution_{#1}}

\newcommand{\loss}{\mathcal{L}}
\newcommand{\out}{y}
\newcommand{\outp}[1]{\out^{#1}}
\newcommand{\outu}[1]{\out_{#1}}
\newcommand{\outpu}[2]{\out_{#1}^{#2}}

\newcommand{\model}{x}
\newcommand{\modelp}[1]{\model^{#1}}
\newcommand{\modelu}[1]{\model_{#1}}
\newcommand{\modelpu}[2]{\modelp{#1}_{#2}}
\newcommand{\gradient}[1]{\nabla \loss(\model; \dataset{#1})}
\newcommand{\gradientp}[2]{\nabla \loss(\modelp{#2}; \datasetu{#1})}
\newcommand{\llr}{\eta}
\newcommand{\lr}{\gamma}
\newcommand{\gradientpSample}[2]{\nabla \loss(\modelp{#1}, #2)}

\newcommand{\edges}{\mathcal{E}}
\newcommand{\graph}{{G}(\users,\edges)}
\newcommand{\graphh}{{G}(\honest, \edges)}
\newcommand{\edgesp}[1]{\mathcal{E}^{#1}}
\newcommand{\graphp}[1]{{G}^{#1}(\users,\edges)}

\newcommand{\graphhpu}[1]{G^{#1}(\honest, \edges)}

\newcommand{\graphclass}[2]{\mathcal{G}_{{#1},{#2}}}

\newcommand{\nbhonest}{n_h}
\newcommand{\nbbyzantine}{n_b}

\newcommand{\R}{\mathbbm{R}}
\newcommand{\localosss}[1]{\mathcal{L}_{#1}}
\newcommand{\weights}{W}
\newcommand{\stacked}{\Theta}

\newcommand{\neighbors}[1]{N(#1)}
\newcommand{\neighborsp}[2]{\neighbors{#1}^{#2}}
\newcommand{\neighborsout}[1]{N_{out}(#1)}
\newcommand{\neighborsin}[1]{N_{in}(#1)}
\newcommand{\neighborsoutp}[2]{\neighborsout{#1}^{#2}}
\newcommand{\neighborsinp}[2]{\neighborsin{#1}^{#2}}
\newcommand{\adversary}{\mathcal{A}}
\newcommand{\byzFrac}{f}
\newcommand{\byzantine}{\mathcal{B}} 
\newcommand{\honest}{\mathcal{H}} 
\newcommand{\correctp}[1]{C({#1})} 
\newcommand{\correcth}[1]{C_h({#1})} 
\newcommand{\corrects}[1]{C_p({#1})} 
\newcommand{\byz}{B} 

\newcommand{\byzThreshold}{b}
\newcommand{\viewSize}{\mathbf{v}}
\newcommand{\msgin}{M}
\newcommand{\msginPush}{\msgin_{push}}
\newcommand{\msginPull}{\msgin_{pull}}
\newcommand{\msginpPush}[1]{\msginPush(#1)}
\newcommand{\msginpPull}[1]{\msginPull(#1)}

\newcommand{\hashf}[2]{rank_{s_{#1}}(#2)}

\newcommand{\initialSet}{I}
\newcommand{\byzForce}{F}
\newcommand{\randomVariable}[1]{#1}
\newcommand{\randomVariableV}[2]{\randomVariable{#1}_{#2}}
\newcommand{\failProb}{\kappa}
\newcommand{\ode}[2]{\dfrac{\mathrm{d}#1}{\mathrm{d}#2}}
\newcommand{\expect}[1]{\mathbb{E}
[#1]}
\newcommand{\history}{h}
\newcommand{\historyp}[1]{h_{#1}}

\newcommand{\cc}[1]{\textcolor{blue}{// #1}}

\section{Introduction}

Gossip Learning~(GL) recently emerged as a promising decentralized learning paradigm~\cite{hegedHus2019gossip}. In GL, users are connected with each other over a unidirectional communication graph. They regularly receive models from their predecessors in the graph; aggregate the received models with their own; perform local steps and send their updated model to their successors in the graph (also referred to as their view).
In contrast to Federated Learning~(FL)\cite{mcmahan2017communication}, this paradigm offers several advantages by removing the dependency on a central entity: it mitigates single-point-of-failure risks, prevents central manipulation, promotes distributed governance, achieves better scalability~\cite{hegedHus2021decentralized}, and even strengthens participant privacy~\cite{cyffers2022muffliato}.

Recently, a dynamic variant of GL has emerged, in which the neighbors of each user change over time. In practice, this accomplished through a Random Peer Sampling~(RPS) protocol~\cite{voulgaris2005cyclon}, which periodically provides fresh samples of the graph to each node. 

Interestingly, Dynamic GL has been shown to enable a graph-size-independent convergence rate~\cite{song2022communication} and to achieve exact averaging~\cite{ying2021exponential}, thus allowing it to converge faster than its static counterpart with drastically sparser graphs (\eg, one honest neighbor can be sufficient~\cite{ying2021exponential}). These properties make Dynamic GL particularly effective in terms of communication costs and scalability. Moreover, recent works indicate that its continuously changing topology reduces the attack surface of curious adversaries~\cite{belal2025inferring, touat2024scrutinizing}.

One relevant yet seemingly orthogonal problem is the resilience of GL to Byzantine attacks. In these attacks, a subset of users (referred to as Byzantine users) may behave arbitrarily and deviate from the learning algorithm. Particularly, they can craft and send poisoned models~\cite{xie2020fall,baruch2019little,DBLP:conf/ndss/KraussKDK24,DBLP:conf/ndss/RiegerKMDS24} to cause model divergence among honest nodes~\cite{guerraoui2024fundamentals}. This problem is challenging due to the limited interpretability of model parameters~\cite{zhang2020interpretable} and is exacerbated by the heterogeneity of training data~\cite{allouah2023fixing}. 
To filter out or bound the impact of such poisoned models, several so-called robust aggregators have been designed for GL~\cite{gaucher2024achieving,Wu2023byzantine,he2022byzantine}. While effective, these solutions do not account for dynamic graphs or nodes that cheat within the RPS.


The problem of Byzantine nodes tampering with the RPS is well established in the distributed systems community. It often takes the form of Byzantine identifiers flooding the network, which results in a disproportionate representation of Byzantine nodes in the neighborhoods of honest nodes(\eg, Hub attacks~\cite{jesi2010secure}, Eclipse attacks~\cite{heilman2015eclipse,singh2006eclipse}). 
As we show in this work, such attacks exacerbate the impact of model poisoning leading to unstable learning and often model divergence.


Byzantine-resilient RPS protocols (BRPS) have been proposed~\cite{auvolat2023basalt,bortnikov2008brahms} to provide random samples with high probability, even in the presence of Byzantine nodes that may, for instance, flood the network. This leaves open the possibility of arbitrary over representations of Byzantine nodes in a local neighborhood, but guarantees it to be temporary. In the applications that originally motivated BRPS, such as reliable message dissemination, this property is not considered a drawback. Indeed, in such applications, ensuring that a message eventually reaches its honest destination is often sufficient. In contrast, GL imposes different requirements. When Byzantine nodes are over-represented in the views of honest nodes over multiple rounds, they can inject poisoned models that not only affect immediate neighbors but also diffuse hop after hop through the network, triggering a domino effect that ultimately leads to the collapse of the learning process. This highlights the inadequacy of existing BRPS protocols for GL: ensuring message dissemination alone is not enough.




To address this gap, there is a need for a BRPS protocol specifically tailored to Dynamic GL, one that fulfills the conditions for robust aggregation to be consistently effective. In particular, robust aggregators such as Clipped Summation~(CS)~\cite{gaucher2024achieving} require accurate control over the proportion of Byzantine nodes in each honest node’s view at every round. Unfortunately, current protocols like BASALT~\cite{auvolat2023basalt} can only constrain this proportion on \textit{average across rounds}. This misalignment undermines the assumptions of robust aggregation techniques and ultimately leads to the divergence of honest nodes' models.

\noindent \textbf{Our contributions.} In this paper, we aim at bridging the gap between BRPS protocols and robust Dynamic GL, two lines of work previously studied in isolation. To this end, we propose \textbf{\system}, a Gossip Learning framework in a non independently and identically (\ie, non-i.i.d) data distribution setting that enables learning over sparse, dynamic graphs in the presence of a Byzantine fraction $\byzFrac$ of nodes, which can simultaneously send poisoned models and corrupt the peer sampling process. \system combines two key components: (i) \textbf{a History-aware Peer Sampling protocol (HaPS)}, which maintains a monotonically expanding set of encountered identifiers to limit the proportion of Byzantine nodes in local neighborhoods. We formally analyze HaPS and show that the time-dependent proportion of Byzantine nodes in local views, $B(t)$, decays exponentially over time. We use this insight to design the second component of \system, (ii)\textbf{ an Adaptive Probabilistic Threshold (APT)}, which leverages $B(t)$ and a Chernoff bound~\cite{chernoff1952measure} to compute adaptive filtering thresholds that guarantee, with high probability, the correctness of the underlying robust aggregator. Finally, we formally prove that \system eventually satisfies the spectral and connectivity conditions (with high prob.) required for robust aggregation, thus guaranteeing convergence under standard assumptions.


Our empirical evaluation demonstrates that \system enables robust Gossip Learning even under strong adversarial conditions, including two of the strongest model poisoning attacks~\cite{xie2020fall,baruch2019little}, various degrees of flooding attacks and up to 30\% of Byzantine presence. We evaluate \system using two robust aggregators, namely Clipped Summation~(CS) and Geometric Trimmed Summation~(GTS)~\cite{gaucher2024achieving}, and compare \system to a state-of-the-art BRPS, namely BASALT~\cite{auvolat2023basalt}. In particular, we show that \system (i) converges with up to 30\% Byzantine nodes, (ii) improves convergence speed through adaptive thresholding, and (iii) enables robust learning under graphs that are up to 9 times less costly communication wise than required by the theory of robust aggregators.

\noindent \textbf{Outline.} The remainder of the paper is structured as follows. Section~\ref{subsec:background} provides the necessary background on GL, robust aggregators and RPS protocols. In Section~\ref{sec:sys}, we present our system model, an overview of \system and detail its algorithm. In Section~\ref{sec:theoritical_analysis}, we outline our assumptions and provide a theoretical analysis of both components of \system, as well as a broader convergence guarantee. We then present our experimental setup in Section~\ref{sec:experiments_granit} before delving into our results in Section~\ref{sec:results}. We then move on to discuss limitations and broader considerations in Section~\ref{sec:discussion} before giving an overview of the related works in Section~\ref{sec:related_work}. Finally, we conclude the paper in Section~\ref{sec:conclusion_granit}.

\section{Preliminaries}
\label{subsec:background}
\subsection{Notation}
For convenience, we summarize in Table~\ref{tab:notation} the main symbols used throughout this work. 
\begin{table}[!htp]
	\caption{Summary of notation.}~\label{tab:notation}
		\centering
		\begin{tabular}{ll}
			\toprule
			\textbf{Symbol} & \textbf{Description} \\
			\midrule
			$\users$ & Set of all nodes\\
			$\honest$, $\byzantine$ & Honest and Byzantine node sets \\
			$\byz$ & Number of Byzantine nodes ($|\byzantine|$) \\
			$\neighborsoutp{i}{t}$ & Outgoing neighbors of node $i$ at time $t$ \\
			$\neighborsinp{i}{t}$ & Incoming neighbors of node $i$ at time $t$ \\
			$\initialSet$ & Initial set of known nodes \\
			$\historyp{i}(t)$ & History of IDs seen by node $i$ up to time $t$ \\
			$\nbusers$ & Total number of nodes \\
			$\viewSize$ & Local view size\\
			$[\viewSize]$ & $\{1,2,\ldots, \viewSize\}$\\
			$\byzFrac$ & Byzantine fraction: $\byz / \nbusers$ \\
			$\hashf{i}{v}$ & Ranking function with seed $s_i$ and input peer $v$ \\
			$\correctp{t}$ & Honest IDs in a node’s history at $t$ \\
			$B(t)$ & Est. local Byzantine ratio at $t$ \\
			$\byzThreshold$ & Byzantine filtering threshold \\
			$b(t)$ & Adaptive threshold at time $t$ \\
			$\failProb$ & Chernoff bound failure prob. \\
			$\sigma$ & Avg. arrival rate of honest IDs \\
			\bottomrule
		\end{tabular}
\end{table}

\subsection{Dynamic Gossip Learning}~\label{subsec:background_gl}
Dynamic Gossip Learning~(GL) is a fully decentralized computing paradigm~\cite{ormandi2011} that enables a set of nodes to collaboratively train a machine learning model through peer-to-peer model exchanges over a dynamic graph. Formally, let $\graphp{1}, \ldots, \graphp{T}$ be a series of time-dependent and undirected graphs, where $\users = \{1,2,\ldots,n\}$ is the set of users and $\edgesp{\round} \subseteq \users \times \users$ denotes the set of communication links at round $\round$. The corresponding adjacency matrix $\adjacency^t$ defines a set of neighbors to gossip with for each node $i$, say $\neighborsp{i}{t}$. Each node $i$ owns local data and uses it to optimize a local model parameterized by $\modelu{i}$. The goal is to find $\model^{*}$ such that:

\begin{equation}\label{eq:uniform_objective}
\model^{*} = \argmin{\model} \frac{1}{\nbusers} \sum_{i = 1}^{\nbusers} \localosss{i}(\model)
\end{equation}
where $\localosss{i}(\model)$ denotes the empirical local loss of node $i$.

To this end, a common approach consists of alternating between gossip averaging steps and local model updates. 
\begin{enumerate}
    \item \textbf{Gossip Step:} Given $\matrice = (\modelu{1},\modelu{2},\ldots, \modelu{\nbusers})^T$, a gossip step can be formulated~\cite{scaman2017optimal,kovalev2020optimal} as:
    \begin{equation}\label{eq:gossip_update}
     \matricep{t+1} = (\identity - \lr \mixing^t) \matricep{t}
    \end{equation}
     where $\lr$ is a communication step and $\mixing^t$ a weighting matrix. For this later, similar to previous work~\cite{gaucher2024unified}, we instantiate the Laplacian of the graph, $\mixing^t = \degree^t - \adjacency^t$, where $\degree^t$ represents the diagonal of degrees matrix. This step corresponds to each node $i$ waking up and pulling the models of their neighbors $\{\modelpu{t}{j} \mid j \in \neighborsp{i}{t}\}$, then aggregating them following $\mixing^t$. 
    \item \textbf{Local training:} Subsequently, each node $i$ performs one or many gradient steps, locally: 
    \begin{equation}\label{eq:local_update_gossip}
        \modelpu{t+1}{i} = \modelpu{t+\frac{1}{2}}{i} - \llr \nabla \loss(\modelpu{t+\frac{1}{2}}{i})
    \end{equation}
    where $\llr$ is a learning rate.
\end{enumerate}

\subsection{Byzantine-Resilient Learning and Robust Aggregators}~\label{subsec:robustagg}
In the presence of a set of Byzantine nodes $\byzantine$, the update rule prescribed by equation~\ref{eq:gossip_update} is not robust and can be diverted to yield any arbitrary value~\cite{blanchard2017machine}. In this scenario, the optimization objective~\ref{eq:uniform_objective} spans over the set of honest nodes only, let it be $\honest$. To achieve it, the simple averaging step is replaced with a robust aggregation that satisfies the definition below. 

\begin{definition}
[$\alpha$-robustness]~\label{def:robustness}
Let $\alpha < 1$. an algorithm aggregating on a graph $\graph$ that contains $|\honest| = \nbhonest$ honest nodes, is said to be $\alpha$-robust if from any initial models $(\modelu{i})_{i \in \honest}$, it enables each honest node to compute $\outu{i}$ such that:  
\begin{equation}
    \frac{1}{\nbhonest} \sum_{i \in \honest} || \outu{i} - \bar{\model}_{\honest} ||^2 
    \le \alpha \frac{1}{\nbhonest} \sum_{i \in \honest} || \modelu{i} - \bar{\model}_{\honest} ||^2
\end{equation}
\end{definition}
By imposing $\alpha < 1$, this notion of robustness forces honest models to move closer to the honest average $\bar{\model}_{\honest}$ (on average) after one aggregation step, which implies that the bias introduced by Byzantine nodes needs to be smaller than the variance reduction between honest models. Note that this property can only be guaranteed for a class of graphs $\Gamma_b$~\cite{gaucher2024unified}, defined as: 
\begin{definition}~\label{def:spectral_condition}
A graph $\graph$ is said to belong to the class of graphs $\Gamma_b$ iff 
\begin{equation}
\Gamma_{b} =\Bigg\{G \bigg|\;\lambda_2(\mixing_\honest) \ge 2b \;\text{and}\; \argmax{i \in \honest}|\neighbors{i} \cap \byzantine| \le b \Bigg\}
\end{equation}
Where $\lambda_2(\mixing_\honest)$ is the smallest non-zero eigenvalue of the Laplacian $\mixing_{\honest}$ of the subgraph induced by $\honest$.
\end{definition}

Intuitively, this definition selects a subset of  sufficiently connected graphs where Byzantine nodes are well spread and their presence in local neighborhoods is bounded.

\noindent \textbf{Robust aggregators.} Several aggregators have been shown to be $\alpha$-robust. In this work, we leverage Geometric Trimmed Summation~(GTS) and Clipped Summation~(CS)~\cite{gaucher2024unified}. At each round $\round$ for each node $i$, both GTS and CS begin by computing neighbor differences sorted by norm:

\begin{enumerate}
    \item For each neighbor $j \in \neighborsp{i}{\round}$, define $y_j^t = \modelpu{\round}{j} \;-\; \modelpu{t}{i} $.
    \item Compute the set of norms: $\{\,||y_j^t|| : j \in \neighborsp{i}{t}\}$. 
    \item Sort the vectors $\{y_j^t\}$ and relabel \textit{s.t.}
    \begin{align*}
      &y^t_{(1)}, y^t_{(2)}, \dots, y^t_{(\viewSize)},
      &
      \text{where}\quad
      ||y^t_{(1)}|| \le ||y^t_{(2)}|| \le \cdots \le ||y^t_{(\viewSize)}||
    \end{align*}  
\end{enumerate}

The GTS and CS are then defined as follows.
 
\begin{definition}
[Geometric Trimmed Sum]~\label{def:gts}
Given an integer trimming parameter $\byzThreshold$,the GTS aggregator discards the  $\byzThreshold$ largest-norm differences:
\begin{equation}
	\modelpu{t+1}{i}
	=
	\modelpu{t}{i}
	 +
	\sum_{r=1}^{\viewSize - \byzThreshold} \,y^t_{(r)}	
\end{equation}
\end{definition}

\begin{definition}[Clipped Sum] 
~\label{def:cs}
Let $\pi = ||y_{(\viewSize- 2 \byzThreshold + 1)}^t ||$ be the clipping threshold. The CS aggregator clips the differences to the norm of the $2\byzThreshold$-th largest vector
\begin{equation}
\modelpu{t+1}{i}=\modelpu{t}{i}+ \sum_{r=1}^{\viewSize} \mathrm{clip}\bigl(y^t_{(r)}, \pi\bigr) 
\end{equation}
where
\begin{equation}
\mathrm{clip}\bigl(y^t_{(r)}, \pi\bigr)  =  y^t_{(r)} \cdot \min \biggl\{1, \frac{\pi}{||y^t_{(r)}||} \biggr\}
\end{equation}

\end{definition}

\subsection{Random Peer Sampling service}\label{subsec:rps}
A Random Peer-Sampling service~(RPS) is a protocol that  periodically provides each node with a random stream of neighbors. Generally, these protocols follow a common workflow: given a fixed view size $\viewSize = |\neighborsoutp{i}{t}|$, a stream of received identifiers $\msginPush$ and a stream of pulled identifiers $\msginPull$, the goal is to compute $\neighborsoutp{i}{t+1}$ = $F(\neighborsoutp{i}{t},\msginPush, \msginPull)$ where the function $F$ returns $\viewSize$ new identifiers. Such a process converges towards a uniform sampling of nodes over the whole graph~\cite{busnel2011uniformity}, yielding both robust dissemination guarantees~\cite{jelasity2004peer} 
and resilience to churn~\cite{allavena2005correctness}. These protocols represent a practical solution to render gossiping graphs dynamic and approximate the faster convergence rates~\cite{ying2021exponential} and the privacy enhancements~\cite{touat2024scrutinizing} of Dynamic GL. However, RPS protocols are vulnerable to Byzantine nodes that flood the network with their IDs, with the objective of over-representing themselves in the neighborhood of honest nodes, referred to as Hub Attacks~\cite{jesi2010secure}, or isolating individual nodes, known as Eclipse Attacks~\cite{heilman2015eclipse}. To account for these behaviors, several Byzantine-Resilient RPS solutions have been proposed. The most prominent approaches are based on the logic of incorporating a local peer selection criterion, which cannot be tampered with by Byzantine nodes~\cite{auvolat2023basalt,bortnikov2008brahms}. Such methods have been shown to prevent honest nodes from getting isolated with high probability. While this is a desirable property, it can be argued that a learning setting requires additionally to bound the proportion of Byzantine nodes in local views. One objective of this work is to bridge the gap between Byzantine-Resilient RPS and Byzantine-Resilient Learning.



\section{The \system approach}
\label{sec:sys}

\subsection{System and Threat Model }~\label{sec:threat-model}
We assume a large distributed system in which nodes can either be honest (a.k.a. correct) or Byzantine (a.k.a. malicious). Byzantine nodes are assumed to act strategically with the explicit objective of preventing learning convergence. In line with standard Byzantine-resilient learning literature~\cite{gaucher2024unified, he2022byzantine,fang2022bridge}, we therefore consider adversaries that can send arbitrary model updates, send distinct models to different neighbors, and collude to infer honest models, enabling stronger untargeted poisoning attacks~\cite{baruch2019little,xie2020fall}. These behaviors capture the most effective known attacks against decentralized learning, as they directly target the optimization dynamics rather than merely reducing participation.
In parallel, we assume Byzantine nodes that actively manipulate the RPS protocol by disproportionately disseminating Byzantine identifiers during peer exchanges, with the goal of amplifying their influence or isolating honest nodes. We model this capability via a parameter $\byzForce$, which denotes the number of times a Byzantine node can inject identifiers into a correct node, within a period during which correct nodes sample only once, following the threat model of~\cite{auvolat2023basalt}.
Importantly, our Byzantine model does not exclude other forms of malicious behavior such as message omission, delayed participation, or passive non-cooperation. These behaviors are inherently captured by the Byzantine abstraction, as they correspond to arbitrary deviations from the protocol. However, in decentralized learning, such behaviors are generally less effective at preventing convergence than active model poisoning or adversarial manipulation of the communication graph. In fact, reducing participation or delaying messages often diminishes the adversary’s influence and may even allow honest nodes to benefit from reduced interference and collaborate more effectively. For this reason, we focus our analysis and evaluation on adversarial behaviors that are known to be the most damaging to learning dynamics. In this sense, our threat model captures Byzantine adversaries that deploy the strongest effective attacks against decentralized learning systems.
As is common in peer sampling literature~\cite{auvolat2023basalt,antonov2023securecyclon,bortnikov2008brahms}, we consider Sybil attacks to be orthogonal, though Section~\ref{sec:discussion} discusses how \system can be extended to handle them. Finally, we tolerate dynamic adversaries, \ie, nodes may switch between honest and Byzantine behavior or adapt their attack strength over time, provided that the total number of Byzantine nodes does not exceed~$\byz$.

In terms of communication, we consider that any node $p$ can send a message to any other node $q$ once $p$ knows $q$’s identity. Moreover, we assume a synchronous network model (\ie, message delays are bounded) and reliable communication channels with signed messages. We note, however, that unreliable channels would only influence our framework if they impact honest nodes more than Byzantine ones. 


\subsection{System Overview}
\system belongs to the Dynamic Gossip Learning framework (See Section~\ref{subsec:background_gl}). Specifically, it assumes a set of nodes $\users$ that aim to collaboratively learn a global model $\model$. To achieve this objective, users train models, exchange them through a series of decentralized communication graphs $\graphp{1}, \graphp{2}, \ldots, \graphp{\round}$, and aggregate received ones. The graphs are generated through a Peer Sampling Protocol. \system aims to solve this problem in the presence of $\byz = \byzFrac \nbusers $  Byzantine adversaries that can act on two layers: (i) the Peer Sampling Protocol, which can enable them to isolate nodes or over-represent themselves in the graph and (ii) the learning model, by sending poisoned models to prevent convergence. To counter these threats, \system incorporates two components: (1) a \textbf{History-aware Peer Sampling~(HaPS)}  that limits the proportion of Byzantine nodes in local views, $B(t)$, after an initial warm-up period and ii) an \textbf{Adaptive Probabilistic Threshold~(APT)} mechanism that uses a worst-case estimate of $B(t)$ to adapt the robust aggregator’s filtering threshold $\byzThreshold$, enabling the aggregation of more models with high probability of correctness. All in all, the conjunction of these two components enable Byzantine-resilience over both dynamic and sparse graphs.

In the following two sections, we provide an overview of the History-aware Peer Sampling and the Adaptive Probabilistic Threshold, before delving in Section~\ref{subsec:algorithm} into the \system Algorithm.


\subsubsection{History-aware Peer Sampling~(HaPS)}
\begin{figure}[!htp]
    \centering
	\begin{tikzpicture}[>=stealth, node distance=20mm, every node/.style={font=\sffamily}, table/.style= {minimum width=1.2cm, minimum height=8mm}, header/.style={table, fill=white, draw}, cell/.style={table, fill=white}]
		
		\begin{scope}[local bounding box=all]
			\node[draw, rectangle, rounded corners, minimum width=20mm, minimum height=10mm, fill=orange!40, align=left] (requests) {Push/Pull \\requests};

		\node[cell, below= of requests, yshift=10mm] (c0) {$s_0$};
		\node[left] (seedlabel) at (c0.west) {seed $[\cdot]$}; 
		\node[cell, right= of c0, xshift=-20mm] (c1)  {};
		\node[cell, right= of c1, xshift=-20mm] (c2) {$s_i$};
		\node[cell, right= of c2, xshift=-20mm] (c3) {};
		\node[cell, right= of c3, xshift=-20mm] (c4) {$s_\viewSize$};

		\node[header, below= of c0, yshift=20mm] (h0) {$n_0$};
		\node[left] (viewlabel) at (h0.west) {view $[\cdot]$};
		\node[header, right= of h0, xshift=-20mm] (h1) {\ldots};
		\node[header, right= of h1, xshift=-20mm] (h2) {$n_i$};
		\node[header, right= of h2, xshift=-20mm] (h3) {\ldots};
		\node[header, right= of h3, xshift=-20mm] (h4) {$n_\viewSize$};			
			
		\node[draw, ellipse, fill=purple!10, minimum width=2cm, minimum height=1cm, right= of requests.east] (bag) {$n_i, n_j, \ldots$};s
			
			\node[above] at (bag.north) {History $\{\cdot\}$};
			
			\draw[decorate,decoration={brace,amplitude=10pt, mirror}]
			(bag.south west) -- (bag.south east)
			node[midway, yshift=-8mm, align=center] (bracelabel) {$B(t)$: proportion of\\ Byzantine neighbors};

			\draw[->] (requests.east) -- (bag.west)
			node[midway, sloped, font=\footnotesize, align=left] (arr) {received \\ identifiers};
			
			\node[below] at (arr.south) {\stepcirc{1}};
			
			\node[below=7mm of h2, align=center] 
			(headertext) {\stepcirc{2} Selected from History\\ to minimize $rank_{s_i}(n_i)$};
			\draw[->] (h2) -- (headertext.north);
			
		\end{scope}

		\node[anchor=north west] (distNode) at (-1cm,-4.5cm) {
			\begin{tikzpicture}
				\begin{axis}[
					width=6cm,
					height=3cm,
					xlabel={$t$},
					ylabel={$B(t)$},
					ymin=0, ymax=1,
					xmin=0, xmax=10,
					axis lines=left,
					tick style={black},
					ytick={0,1},
					smooth,
					]
					\addplot[blue, thick] {exp(-0.3*x)};
				    \addplot[red, dashed, thick] coordinates {(0,0.2) (10,0.2)};

				\end{axis}
	    \node[red, left] at (0, 0.5) {$f$}; 
				
		\end{tikzpicture}};

		\node[align=left] at (2.5,-7.5) (th) {\stepcirc{3} Theoretical\\exponential decay};

		\node[draw,dashed,rounded corners,fit=(requests)(h0)(h1)(h2)(h3)(h4)(c0)(c1)(c2)(c3)(c4)(bag)(headertext)(bracelabel)(viewlabel)(seedlabel)(distNode)(th),
		inner sep=5mm,
		fill=gray!20, fill opacity=0.3
		] (haps) {};
		
	\end{tikzpicture}
	\caption{High-level overview of the History-aware Peer Sampling and its Byzantine exponential decay property.\label{fig:haps}}
\end{figure}
Figure~\ref{fig:haps} illustrates the steps of HaPS and its impact on the Byzantine presence within local neighborhoods. As a Peer Sampling Service, the first objective of HaPS is to enable every node $i$ to discover new neighbors. Nodes continuously exchange their local views through stochastic push and pull requests. Through these exchanges, they collect identifiers that accumulate into the node's history (phase \smallstepcirc{1}). We denote this history by $\historyp{i}(t)$; it is a permanently growing set of potential neighbors. 
To incorporate a node from the history into the local view $\neighborsoutp{i}{t}$ at round $\round$, $i$ uses a ranking-based selection rule. Specifically, for each slot $j \in [\viewSize]$, a seed  $s_j$ defines a hash function $\hashf{j}{\cdot}$.
Node $i$ selects $p^* = \argmin{p \in \historyp{i}(t)} \hashf{j}{p}$ for each slot $j$ (phase \smallstepcirc{2}). In practice, we implement $\hashf{j}{\cdot}$ using a uniform hash function $h$, \ie, $\hashf{j}{p}= h(<s_j,p>)$. This enables $i$ to uniformly sample from its history. Since fixed seeds would cause the overlay graph to converge to a static structure, the seeds are refreshed every 
$\pi$ rounds. In our experiments, we use continuously refreshed seeds (\ie, $\pi=1$). 

The second objective of HaPS is to ensure that the proportion of Byzantine nodes in the history of a node $i$, denoted $B(t)$, remains close to their global fraction $\byzFrac$, \ie, $|\byzFrac - B(t)| < \epsilon$. This guarantees that honest nodes are not surrounded by Byzantine ones and therefore still receive a majority of correct models per round. In HaPS, this property holds by design under the assumption that each node initially knows at least one honest node. Combined with the stateful nature of the history, this assumption triggers a cascading effect: knowing a single honest node eventually leads to the propagation of honest identifiers throughout the network. Consequently, after a finite number of rounds, the proportion of Byzantine nodes in each history, $B(t)$, is guaranteed to converge toward its global fraction $\byzFrac$. As we show in Section~\ref{subsec:haps_analysis}, even in a worst-case initialization where every $\historyp{i}(0)$ contains all Byzantine nodes plus a single honest node, $B(t)$  still converges exponentially to $f$ (phase \smallstepcirc{3}). In fact, considering this worst-case scenario, HaPS enables both the estimation of $B(t)$ and the characterization of its exponential convergence toward $\byzFrac$.

\subsubsection{Adaptive Probabilistic Threshold~(APT)}
Robust aggregators are often based on a filtering threshold that specifies the $O(\byzThreshold)$ models to be treated as Byzantine (\eg, filtered-out, clipped, \ldots). In static settings, $\byzThreshold$ is either assumed known or chosen conservatively. However, this approach is unfeasible in a dynamic setting, as discussed in Section~\ref{subsec:robustagg}. HaPS mitigates this issue by estimating and bounding $B(t)$ in a worst-case scenario (see Theorem\eqref{thm:convergence-c}). Figure~\ref{fig:apt_overview} illustrates how this estimate is leveraged to design an Adaptive Probabilistic Threshold that dynamically adjusts the robust aggregation. First, we note that under a mean-field assumption the expected number of Byzantine nodes in a local view equals to $\viewSize \cdot B(t)$. Let $\failProb$ denote the probability that the number of Byzantine nodes exceeds this expectation by a factor of $(1 + \delta)$. Using a Chernoff bound, we derive a threshold $\byzThreshold(t)$ s.t the probability of effectively exceeding it is smaller than $\failProb$ (step~\smallstepcirc{1}). In other terms, $\byzThreshold(t)$ ensures that, with high probability, the number of Byzantine nodes does not exceed  $(1+\delta) \cdot \viewSize \cdot B(t)$. In a second phase (step \smallstepcirc{2}), $\byzThreshold(t)$ is fed to the robust aggregator to filter (or clip) $\byzThreshold(t)$ models and aggregate the remaining ones (step \smallstepcirc{3}). Compared to a static $\byzThreshold$, this adaptive approach allows the robust aggregation to adjust to the learning stage: early stages require larger thresholds, but as nodes discover more correct neighbors and their neighborhoods become less dominated by Byzantine nodes, smaller thresholds suffice, enabling aggregation over larger numbers of models and thus, faster convergence. The detailed computation of $\byzThreshold(t)$ and its theoretical guarantees are deferred to Section~\ref{subsec:apt_analysis}.

\begin{figure}
\centering
	\begin{tikzpicture}[>=Stealth, node distance=20mm, every node/.style={font=\sffamily}]
		
		\node[draw, circle, minimum size=3mm] (x1) at (-3,-3) {$\modelpu{t}{1}$};
		
		\node[draw, circle, minimum size=3mm, below=8mm of x1] (x2) {$\modelpu{t}{2}$};
		
		\node[below=1mm of x2] (dots) {$\vdots$};
		
		\node[draw, circle, minimum size=6mm, below=4mm of dots] (xn) {$\modelpu{t}{n}$};


		\node[draw, rectangle, rounded corners, minimum width=10mm, minimum height=5mm, right=12mm of x2, fill=blue!40] (agg) {Filtering};
		
		\node[below] at (agg.south) {\stepcirc{2}};
		\draw[->] (x1) -- (agg.west);
		\draw[->] (x2) -- (agg.west);
		\draw[->] (xn) -- (agg.west);

		\begin{scope}
			\node[draw, rectangle, rounded corners, minimum width=10mm, minimum height=8mm, yshift=35mm] (B) {$B(t)$};
			

			\begin{scope}[local bounding box=distplot, shift={(B.east)}, xshift=-2.8cm, yshift=-4cm]
				
				\draw[->] (0,0) -- (4.6,0) node[below right]{$X^t$};
				\draw (0,0) -- (0,2.4) node[left]{$\Pr$};

				\draw[very thick, smooth, samples=100, domain=0.3:4.2]
				plot(\x,{1.8*exp(-((\x-2)^2)/0.55)});
				
				\foreach \x/\h in {0.9/0.35,1.3/0.65,1.8/1.3,2.2/1.4,2.7/0.9,3.2/0.45}{
					\draw[thick] (\x,0) -- (\x,\h);
					\filldraw[black] (\x,\h) circle (1pt);
				}
				
				\draw[dashed] (1.9,0) -- (1.9,2) node[above]{$\viewSize \cdot B(t)$};
				
				\coordinate (x) at (2.8,0);
				\draw[thick, blue!60] (x) -- ++(0,1.2)
				node[right] (tailnode) {\footnotesize $(1+\delta) \cdot \viewSize \cdot B(t)$};

				\fill[red!70!black]
				(x)
				-- plot[smooth, samples=100, domain=2.8:4.2]
				(\x,{1.8*exp(-((\x-2)^2)/0.55)})
				-- (4.2,0)
				-- cycle;

				\node[anchor=west, red!70!black] at (3.2,0.5)
				{\footnotesize $Surface\le \kappa$};
				
			\end{scope}
			\draw[->] (B) -- (distplot.north);
    		\node[below] at (distplot.south west) {\stepcirc{1}};
		
		\end{scope}
		
		\node[draw,dashed,rounded corners,fit=(distplot),inner sep=7.5mm, fill=gray!20, fill opacity=0.3] (apt) {};
		
		\draw[->, thick, blue!60, bend right=20] (x) to (agg.north)
		node[above=2cm, font=\footnotesize] {$\byzThreshold(t)$};
		\node[draw, circle, minimum size=3mm, fill=green!30, right= 3cm of x1] (y1){$\modelpu{t}{1}$};
		\node[draw, circle, minimum size=3mm, fill=green!30, below=8mm of y1] (y2) {$\modelpu{t}{2}$};
		
		\node[below=1mm of y2] (dots2) {$\vdots$};
		
		\node[draw, circle, minimum size=3mm, fill=green!30, below=2mm of dots2] (yn) {$\modelpu{t}{n - \lfloor \byzThreshold(t)\rfloor + 1}$};
		
		\draw[->] (agg.east) -- (y1);
		\draw[->] (agg.east) -- (y2);
		\draw[->] (agg.east) -- (yn);
		
		\node[draw, rectangle, rounded corners, minimum width=10mm, minimum height=5mm, right=12mm of y2, fill=orange!50] (avg) {Aggregation};

		\draw [->] (y1) -- (avg.west);
		\draw [->] (y2) -- (avg.west);
		\draw [->] (yn) -- (avg.west);

		\draw[->] (x1) -- (agg.west) node[pos=0.3, right, sloped, font=\footnotesize, align=left] {received\\models};
		
		\draw[->] (agg.east) -- (y1) node[pos=0.2, right, sloped, align=left, font=\footnotesize] {filtered\\models};
		
		\node[below] at (avg.south) {\stepcirc{3}};
		
		\node[draw,dashed,rounded corners,fit=(agg)(y1)(yn)(avg),inner sep=3mm, fill=gray!20,  fill opacity=0.3] (robagg) {};

	\end{tikzpicture}
	
	\caption{Illustration of the Adaptive Probabilistic Threshold and its role in the Robust Aggregation process.~\label{fig:apt_overview}}
\end{figure}

\subsection{The \system algorithm}~\label{subsec:algorithm}
Algorithm~\ref{alg:sys} summarizes the main steps of \system, which combines the History-aware Peer Sampling with the Adaptive Probabilistic Threshold mechanism. 

Lines~\ref{alg:sys-push}-~\ref{alg:sys-merge} implement the push-and-pull exchange of identifiers that enable nodes to discover new neighbors. Newly discovered identifiers are inserted in a node's history (line~\ref{alg:sys-history-update}). HaPS then leverages this history to sample nodes using uniform ranking hash functions applied in a round-robin manner (lines~\ref{alg:sys-view-update1} and \ref{alg:sys-view-update2}). 

In a second phase, each node estimates the proportion of correct nodes in its neighborhood, $\correctp{t}$, using only its initial number of known honest nodes $\correctp{0}$, the global number of honest nodes $|\honest|$ and the exchange rate $\sigma$ (line~\ref{alg:sys-ct}). This approximation relies on the stateful nature of HaPS; we defer the derivation to Section~\ref{subsec:haps_analysis}. Under the worst-case assumption that every node already received all Byzantine identifiers (see Section~\ref{subsec:assumptions}), nodes then estimate the expected local proportion of Byzantine neighbors $B(t)$ (line~\ref{alg:sys-bt}). Finally, given $B(t)$ and a user-chosen failure probability $\kappa$, nodes then compute $\delta$ and derive $\byzThreshold(t)$, the adaptive filtering threshold (line~\ref{alg:sys-threshold}) fed into the robust aggregation. The construction of $\byzThreshold(t)$ and its probabilistic guarantees are detailed in Section~\ref{subsec:apt_analysis} .



\SetKwInOut{parameters}{parameters}
\begin{algorithm}[!htb]
\caption{\system}
\label{alg:sys}

\parameters{Initial view $\neighborsoutp{i}{0}$ from $\initialSet$, view size $\viewSize$, 
	       ranking functions $\{\hashf{i}{v}\}_{v=1}^{\viewSize}$, total number of rounds $T$, 
	       number of Byzantine nodes $\byz$, number of honest nodes $|\honest|$, 
	       failure probability $\failProb$}

\cc{Initialization} \\
$\historyp{i}(0) \gets \initialSet$\;
$\neighborsoutp{i}{0} \gets$ select best $\viewSize$ IDs from $\historyp{i}(0)$ using $\{\hashf{i}{v}\}$\;

\For{$t = 1$ \KwTo $T$}{
	
	    \cc{Push, Pull interactions and history update}\\
	    Select $j \in \neighborsoutp{i}{t-1}$ at random;
	    Send $\neighborsoutp{i}{t-1}$ to $j$\;\label{alg:sys-push} 
	    
	    Select $k \in \neighborsoutp{i}{t-1}$ at random;
	    Request $\neighborsoutp{k}{t-1}$\;\label{alg:sys-pull} 
	    
	    Receive $\msginpPush{i}$ and $\msginpPull{i}$\;\label{alg:sys-merge}
	    
	    $\historyp{i}(t) \gets \historyp{i}(t-1) \cup \msginpPush{i} \cup \msginpPull{i}$\;\label{alg:sys-history-update} 
	    
	    \cc{View refresh}\\
	    \For{$v = 1$ \KwTo $\viewSize$}{
		      \quad \quad $p^* \gets \argmin{p \in \historyp{i}(t)} \hashf{v}{p}$\;\label{alg:sys-view-update1}
		      \quad \quad $\neighborsoutp{i}{t}[v] \gets p^*$\;\label{alg:sys-view-update2}
		    }
	    \cc{Estimate $C(t)$ and $B(t)$}\\
	    $\correctp{t} \gets |\honest| - (|\honest| - \correctp{0}) \cdot e^{-\frac{\sigma}{|\honest|} t}$\;\label{alg:sys-ct}
	    $B(t) \gets \frac{\byz}{\byz + \correctp{t}}$\;\label{alg:sys-bt}
	    
	    \cc{Compute adaptive threshold $\byzThreshold(t)$}\\
	    $\delta \gets \frac{-\ln(\failProb) 
		                      + \sqrt{\ln(\failProb)^2 - 8 \cdot \viewSize \cdot B(t) \cdot \ln(\failProb)}}
	                      {2 \cdot \viewSize \cdot B(t)}$\;\label{alg:sys-delta}
	
	    $b(t) \gets \min\bigl((1 + \delta) \cdot \viewSize \cdot B(t) , \viewSize - 1\bigr)$\;\label{alg:sys-threshold}
	    
	    }
\end{algorithm}

\section{Worst-case Theoretical Analysis}~\label{sec:theoritical_analysis}
In this section, we use a theoretical continuous model to prove an exponential decay of $B(t)$, the local proportion of Byzantine nodes, under HaPS, and describe the probabilistic guarantees of APT. We further demonstrate that after a warm-up period, \system enters a regime where it generates graphs that satisfy the spectral connectivity assumptions of $\alpha$-robustness with high probability. Together, these results guarantee theoretical convergence of \system under standard optimization and data distribution assumptions~\cite{gaucher2024unified}.

\subsection{Parameters and Assumptions}~\label{subsec:assumptions}
\subsubsection{Scenario Parameters}
We note $\correctp{t}$ the expected number of unique identifiers in the history at round t. $\sigma$ denotes the expected arrival rate of honest identifiers through push/pull requests. Let $\randomVariable{Z}_t$ be the number of Byzantine nodes in a local view at round $t$.
\subsubsection{Assumptions}
We operate our analysis under the following assumptions:
\begin{itemize}
	\item \textbf{Uniform ranking function.}
	The ranking hash function used for sampling views is assumed to be perfectly uniform over its output space.
	\item \textbf{Bootstrapping set.}
	Each node joining the system receives an initial random set of identifiers $\initialSet$, sampled i.i.d.\ as in~\cite{auvolat2023basalt}, from which it selects $\viewSize$ neighbors. We assume that $\initialSet$ contains at least one honest identifier.
	\item \textbf{Worst-case initialization of histories.}
	The initial local histories of honest nodes contain all Byzantine identifiers: 
	$\forall i \in \honest: \byzantine \subset \historyp{i}(0)$
	\item \textbf{Mean-field approximation.}
	We analyze the dynamics under a mean-field assumption, treating node interactions as independent and identically distributed across the population. Moreover, all quantities are evaluated in expectation, as these tend to concentrate around their means with high probability.
\end{itemize}

\subsection{HaPS: Convergence of Byzantine Proportion}~\label{subsec:haps_analysis}
In this section, we aim to study the evolution of the expected local Byzantine ratio $B(t)$ and derive a conservative upper bound for it. To this end, we characterize the convergence of the known honest nodes towards $\honest$ in the following theorem.
 
\begin{theorem}[Exponential Convergence of Known-Honest IDs]
	\label{thm:convergence-c}
	Let $\correctp{t}$ denote the number of unique honest identifiers known to a node at time $t$, and assume that correct identifiers arrive at an overall rate \(\sigma\). We then have:
	\[
	\correctp{t} = |\honest| - \left(|\honest| - \correctp{0}\right) \exp\left(-\frac{\sigma}{|\honest|} t\right)
	\]
\end{theorem}

\begin{corollary}
	The expected time-dependent fraction of
	Byzantine nodes in local views is bounded by:
	\[
	B(t) = \frac{\byz}{\byz + |\honest| - \left(|\honest| - \correctp{0}\right) \exp\left(-\frac{\sigma}{|\honest|} t\right)}
	\]
\end{corollary}

This result shows that $B(t)$ decreases exponentially and converges to $\byzFrac$, the global Byzantine fraction, even under adversaries with unlimited communication capabilities. We prove this theorem in Appendix~\ref{appendix:ode-derivation}.

\subsection{Adaptive Probabilistic Threshold Guarantee~(APT)}~\label{subsec:apt_analysis}
\begin{lemma}[APT Filtering Guarantee]
	For any $\failProb \in (0,1)$, following Algorithm~\ref{alg:sys}, a node in APT selects the smallest $\delta > 0$ and uses a filtering threshold \[
	\byzThreshold(t) = (1 + \delta) \cdot \viewSize \cdot B(t) \] such that:
	\[
	P\left(\randomVariable{Z}_t \geq \byzThreshold(t) \right) \leq \failProb
	\]
	Where $\randomVariable{Z}_t$ is the r.v capturing the number of Byzantine nodes in local neighborhoods. In other terms, this ensures that, with probability at least $1 - \failProb$, the number of Byzantine nodes in the local view is smaller than the filtering threshold $b(t)$ selected by Algorithm~\ref{alg:sys}.
\end{lemma}

\begin{proof}
Given our assumptions, each slot is occupied by a Byzantine node independently with probability at most $B(t)$, and $\expect{\randomVariable{Z}_t} = \viewSize \cdot B(t)$. Since $\randomVariable{Z}_t$ is a sum of independent Bernoulli variables with success probability at most $B(t)$, then by the Chernoff upper-tail bound~\cite{mitzenmacher2017probability}, for any $\delta>0$:
\begin{equation}
	P\left(\randomVariable{Z}_t \geq (1 + \delta) \cdot  \viewSize \cdot   B(t)\right)
	\leq \exp\left(- \frac{\delta^2 \cdot  \viewSize \cdot  B(t)}{\delta + 2} \right)
\end{equation}

Setting the right-hand side equal to $\kappa$ and solving for $\delta$ leads each node to compute based on current value of $B(t)$:
\begin{equation}
	\delta = \frac{-\ln(\failProb) + \sqrt{\ln(\failProb)^2 - 8 \viewSize \cdot B(t) \cdot \ln(\failProb)}}{2 \cdot \viewSize \cdot B(t)}
\end{equation}
\end{proof}

\noindent \textbf{Practical Consideration.}
In rare cases, the computed threshold $\byzThreshold(t)$ may exceed the view size $\viewSize$, particularly if the failure probability $\failProb$ is chosen to be extremely small, or if the estimated $B(t)$ is loose in early rounds. To account for this, we compute:
\[
\byzThreshold(t) \gets \min\bigl(\byzThreshold(t), \viewSize - 1\bigr).
\]
This corresponds to discarding all but one model in the worst case. While this fallback is not theoretically backed by the Chernoff bound, it remains safe in practice. Indeed, given that node isolation is unlikely and at least one honest model is expected in most views, the model retained still has a high probability of being honest.

\subsection{Robustness of \system}
In \system, (honest) communications occur over a sequence of graphs
$\graphp{1}, \ldots, \graphp{T}$. Under the worst-case initial conditions considered by APT, the initial honest subgraphs fail to satisfy the spectral connectivity properties required for $\alpha$-robustness. Let us investigate if \system eventually generates graphs that do satisfy such conditions. At each round $t$, the honest subgraph $\graphhpu{t}$ is generated by the protocol dynamics and can be modeled as a random graph in the class
\begin{equation}
\begin{aligned}
\graphhpu{t} \sim \mathcal{G}(n_h, p_t),
\qquad p_t = 1 - B(t),
\end{aligned}
\end{equation}

where $n_h$ denotes the number of honest nodes and $p_t$ is the probability of sampling an honest node. This representation allows us to leverage classical spectral results for random graphs~\cite{feige2005spectral}. Specifically, by applying theorem~\ref{thm:er_spectral_appendix} to the honest subgraph $\graphhpu{t}$, we obtain (see Appendix~\ref{app:derivation} for derivation)
\begin{equation}
\begin{aligned}
\lambda_2(\mixing_\honest^t) \ge \viewSize (1 - B(t)) - c'\sqrt{n\bigl(1 - B(t)\bigr)} ,
\end{aligned}
\end{equation}

with high probability, where $\viewSize(1-B(t))$ is the expected honest degree induced by \system.

We now relate this emergent spectral connectivity to the requirements of $\alpha$-robustness. Rather than reasoning over the sequence of graphs generated by \system jointly, we take a conservative approach and verify that each graphs independently belongs to $\Gamma_b(t)$, a natural generalization of Definition~\ref{def:spectral_condition}, where the Byzantine neighbor bound $b$ is replaced by its time-dependent counterpart $b(t)$. Crucially, this generalization preserves the theoretical guarantees of $\alpha$-robustness, since the $\Gamma_b(t)$ conditions are verified at each graphs with high probability. Formally, we study when the graphs generated by \system satisfy:
\begin{equation}
\begin{aligned}
\lambda_2(\mixing_\honest^t) \;\ge\; 2 b(t),
\qquad\text{with}\qquad
b(t) = \viewSize(1+\delta)B(t),
\end{aligned}
\end{equation}
where $\delta \ge 0$ is the confidence parameter introduced by APT.


\begin{theorem}[Robust connectivity]
\label{thm:robustness_threshold}
Under the assumptions above, a sufficient condition for robustness of \system at round $t$ is
\[
B(t)
\;\le\;
\frac{1 - \frac{c'}{\viewSize}\sqrt{n\bigl(1 - B(t)\bigr)}}{3 + 2\delta}.
\]

In particular, neglecting the lower-order spectral correction term
$\frac{c'}{\viewSize}\sqrt{n(1 - B(t))}$, this condition reduces to
\[
B(t) < \frac{1}{3 + 2\delta}.
\]
\end{theorem}


Intuitively, Theorem \ref{thm:robustness_threshold} indicates that, the honest sub-graphs generated by \system become sufficiently connected to have robust aggregation, with high probability, as long as the proportion of Byzantine nodes at any given round is strictly smaller than $1/3$, with this fraction decreasing as the confidence parameter $\delta$ increases. This result recovers a classical limit of Byzantine resilient learning~\cite{blanchard2017machine} as a special case ($\delta=0$). It is clear that early graphs generated by \system will not meet this condition but once $B(t)$ decreases sufficiently, which we have shown to be exponential, the system will enter a regime where the generated graphs are highly likely to align with $\alpha$-robustness. Obtaining a closed form for the the number of rounds needed to generate such graphs is non-trivial. However, we can characterize what controls this number. Indeed, Lemma~\ref{lem:robustness_convergence} illustrates that there are two main factors: the honest network size $n_h$ and the honest identifier arrival rate $\sigma$. This indicates that for larger networks, honest nodes can compensate by having more push and pull requests, thus, receiving more honest identifiers and accelerating convergence towards the "good graphs" regime.

\begin{lemma}[Convergence to Robust Connectivity]
\label{lem:robustness_convergence}
After a finite number of rounds, the honest sub-graph satisfies the spectral connectivity conditions required for $\alpha$-robustness. Specifically, it suffices to have:
\[
T \geq \tau \cdot \frac{n_h}{\sigma} \ln \frac{n_h}{n_h - \correctp{0}}
\]
for a constant $\tau > 0$ depending on $B$, $\delta$, and $\correctp{0}$, so that $T = O\!\left(\dfrac{n_h}{\sigma}\right)$. In particular, larger networks require more rounds, but this can be compensated by increasing the identifier exchange rate $\sigma$.
\end{lemma}

All in all, these results demonstrate that while individual graphs can fail to satisfy the robustness conditions, a careful choice of $\delta$ enables \system to enter a regime graphs satisfying the robustness conditions are generated with high probability. Moreover, the speed at which this regime is achieved can be controlled through $\sigma$. Once achieved, \system aligns with $\alpha$-robustness over static graphs (with high probability), which, when combined with standard assumptions w.r.t the optimization problem and data distribution, guarantees theoretical convergence~\cite{gaucher2024unified}.

\section{Evaluation}
\label{sec:experiments_granit}
In this section, we evaluate \system under various adversarial conditions and quantify its resilience under different robust aggregators and compare it against a state-of-the-art Byzantine-resilient peer-sampling protocols. The rest of the section is structured as follows. We first introduce the experimental environment, including datasets, learning models, implemented attacks, and robust aggregators. We then detail our evaluation metrics, followed by a description of competitors. Finally, we present our research questions and the corresponding empirical results. 
\subsection{Evaluation Environment}
The gossiping logic of \system is implemented using Gossipy, a Gossip Learning Python {library\footnote{\url{https://github.com/makgyver/gossipy/}}}. We extend Gossipy to dynamic graphs and adversarial settings. Model training and experimentation are conducted using PyTorch~\cite{imambi2021pytorch}. Experiments were executed on an 11th Gen Intel® Core™ i9-11950H CPU at 2.60GHz with an NVIDIA RTX A5000 Mobile GPU. 

\subsection{Datasets and models}
We evaluate \system on image and tabular datasets, namely, MNIST~\cite{lecun1998mnist} and Purchase100~\cite{shokri2017membership}. For the earlier, we train a CNN comprising two convolutional layers followed by two fully connected layers, similarly to prior work~\cite{gaucher2024unified}. For Purchase100, we follow the architecture of Nasr et al.~\cite{nasr2019comprehensive}, consisting in a 4-layer fully-connected neural network. In both cases, the user data distributions are heterogeneous and generated using a Dirichlet distribution~\cite{li2022federated} with parameter $\beta = 5$.

\subsection{Untargeted Poisoning Attacks}
We implement the two most effective attacks from the literature, both of which have been shown to succeed and exploit the heterogeneity of the system (a setting known to be particularly challenging to defend~\cite{allouah2023fixing}). Namely, Fall Of Empires~(FOE)~\cite{xie2020fall} and Little is Enough~(ALIE)~\cite{baruch2019little}. 
These attacks involve sending $\modelpu{\round}{i} + \zeta_i a_i^\round$ to the victim $i$ where $\zeta_i$ is a scaling factor optimized via linear search and $a_i^\round$ is the attack direction.
\begin{itemize}
    \item \textbf{FOE}: Sets $a^t = - \modelpu{\round}{\honest}$, which is shown to be sufficient to circumvent several defense mechanisms with reasonable stealth.
    \item \textbf{ALIE}: Computes the average honest model $\modelpu{\round}{\honest}$ and $\sigma^{\round}$, its coordinate-wise standard deviation. Then, it sets $a^t= \sigma^{\round}$.
\end{itemize}

\subsection{Flooding Attack}
In addition to the poisoning attacks, we consider a flooding attack wherein Byzantine nodes disproportionately saturate the network with their identifiers. In each round, a honest node sends its current view to a single randomly chosen neighbor. In contrast, a Byzantine node sends $\viewSize$ Byzantine identifiers to $\byzForce$ honest nodes in its neighborhood.
In the strongest variant of the attack, denoted \(\byzForce = \infty\), each Byzantine node reaches out to all neighboring honest nodes.
Additionally, when responding to a pull request, Byzantine nodes systematically return $\viewSize$ Byzantine identifiers. 

\subsection{Robust Aggregators}
Our selection of robust aggregators was guided by their demonstrated effectiveness in fully decentralized learning settings. Specifically, we chose Clipped Summation (CS) and Geometric Trimmed Summation (GTS) (see Definitions~\ref{def:cs} and~\ref{def:gts}). For completeness, we also include the classical Coordinate Wise Trimmed Mean (CWTM)~\cite{yin2018byzantine} as a baseline. Although CWTM is not tailored for fully decentralized learning, it remains a widely used reference point in robust aggregation.

\subsection{Competitors}
We compare \system with \textbf{BASALT~\cite{auvolat2023basalt}}, a state-of-the-art Byzantine-resilient RPS protocol, assigning uniform ranking hash functions to slots in each node’s view and periodically returning fresh samples. Since BASALT focuses on the set of outgoing neighbors of a node, we adapt it to the learning setting by pulling models from this set. This ensures fairness, as BASALT does not provide explicit guarantees regarding the incoming neighborhoods.

\subsection{Metrics} We leverage the following metrics to quantify Byzantine presence and model performance:
\begin{itemize}
    \item \textbf{F1-Score}: a standard classification performance metric. We report an average over all honest nodes. The standard deviation is consistently low (\eg, $\pm$0.015), indicating that the average is a reliable summary of performance across nodes.
    \item \textbf{Optimal}: For comparison, we report the optimal F1-Score that can be achieved by our models over the considered datasets in the absence of any adversarial behavior.
    \item \textbf{$\byzFrac$}: When reported, it serves as a baseline between the ideal setting, where Byzantine nodes are uniformly distributed in the graph, and the observed proportion.
    \item \textbf{$\byzFrac_{in}$}: The average Byzantine fraction in local views.
    \item \textbf{$B(t)$}: we report the higher bound approximation on $\byzFrac_{in}$ to quantify the gap between our model and the empirical observation.


\end{itemize}

\subsection{Evaluation Hyper-parameters}
The main hyperparameters and experimental settings used throughout our evaluation are summarized in Table~\ref{tab:exp_params}.
\begin{table}[h]
\small
\centering
\caption{Experimental Setting Hyperparameters.}
\label{tab:exp_params}
\begin{tabular}{ll}
\toprule
\textbf{Parameter} & \textbf{Value} \\
\midrule
Learning rate $\llr$ & $0.01$ \\
Momentum & $0.9$ \\
Batch size & $32$~(MNIST) / $256$~(Purchase100) \\
Number of nodes ($\nbusers$) & $300$ \\
View size ($\viewSize$) & $20$ \\
Bootstrap size ($\initialSet$) & $30$, $60$ \\
Seeds refreshed per interval & $10$ \\
$\rho$ (BASALT) & $0.25$ \\
Seed refresh interval (BASALT) & $\viewSize / \rho$ steps \\
Byzantine node ratio & $0.1$, $0.3$ \\
Byzantine Force ($\byzForce$) & $1$, $2$, $\infty$ \\
\bottomrule
\end{tabular}
\end{table}

\section{Experimental Results}
\label{sec:results}
In this work, we aim at answering the following research questions:
\begin{itemize}
    \item \textbf{RQ1:} Can \system effectively mitigate the combination of Byzantine flooding and model poisoning? and How does it compare to BASALT?
    \item \textbf{RQ2:}
    How does \system's connectivity and communication efficacy compare to the requirements of standard theory on robust GL? 
    \item \textbf{RQ3:} How do different robust aggregators perform under \system? 
    \item \textbf{RQ4:} How accurately does the analytical approximation $B(t)$ estimate the empirical Byzantine fraction in local views ($\byzFrac_{in}$), under varying attack forces $\byzForce$?
    \item \textbf{RQ5:} What is the impact of churn on \system? 
\end{itemize}
\subsection{\system performance and comparative analysis with BASALT (RQ1)}
\begin{figure*}[!tb]
  \centering
  \includegraphics[width=0.8\textwidth]{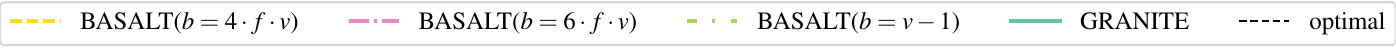}
  \begin{minipage}{\textwidth}
  \centering
  \begin{subfigure}[b]{0.23\textwidth}
    \centering
    \includegraphics[width=\textwidth]{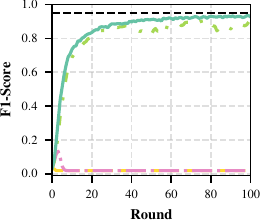}
    \caption{FOE ($\byzFrac=0.1$)}
    \label{subfig:CG_FOE_f1}
  \end{subfigure}
  \begin{subfigure}[b]{0.23\textwidth}
    \centering
    \includegraphics[width=\textwidth]{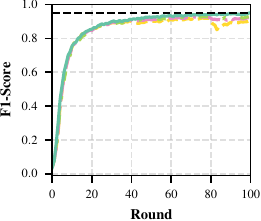}
    \caption{ALIE ($\byzFrac=0.1$)}
    \label{subfig:CG_ALIE_f1}
  \end{subfigure}
  \begin{subfigure}[b]{0.23\textwidth}
    \centering
    \includegraphics[width=\textwidth]{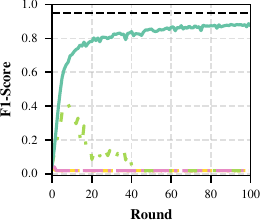}
    \caption{FOE ($\byzFrac=0.3$)}
    \label{subfig:CG_FOE_f3}
  \end{subfigure}
  \begin{subfigure}[b]{0.23\textwidth}
    \centering
    \includegraphics[width=\textwidth]{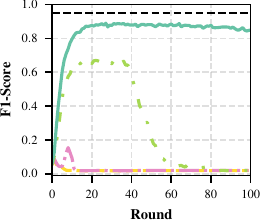}
    \caption{ALIE ($\byzFrac=0.3$)}
    \label{subfig:CG_ALIE_f3}
  \end{subfigure}
  \end{minipage}
  \caption{F1-Score comparison between \system and three flavors of BASALT under $\byzForce=2$ and CS aggregator over MNIST dataset.}
  \label{fig:MNIST_CG}
\end{figure*}

\begin{figure*}
  \centering
  \includegraphics[width=0.8\textwidth]{Plots/outer_legend_cropped.pdf}
  \begin{minipage}{\textwidth}
  \centering
  \begin{subfigure}[b]{0.23\textwidth}
    \centering
    \includegraphics[width=\textwidth]{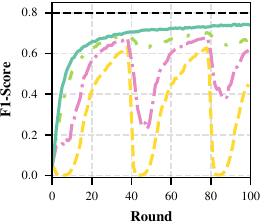}
    \caption{FOE ($\byzFrac=0.1$)}
    \label{subfig:Purchase_CG_FOE_f1}
  \end{subfigure}
  \begin{subfigure}[b]{0.23\textwidth}
    \centering
    \includegraphics[width=\textwidth]{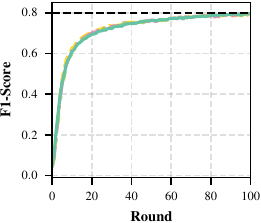}
    \caption{ALIE ($\byzFrac=0.1$)}
    \label{subfig:Purchase_CG_ALIE_f1}
  \end{subfigure}
  \begin{subfigure}[b]{0.23\textwidth}
    \centering
    \includegraphics[width=\textwidth]{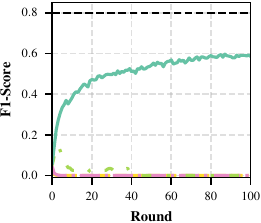}
    \caption{FOE ($\byzFrac=0.3$)}
    \label{subfig:Purchase_CG_FOE_f3}
  \end{subfigure}
  \begin{subfigure}[b]{0.23\textwidth}
    \centering
    \includegraphics[width=\textwidth]{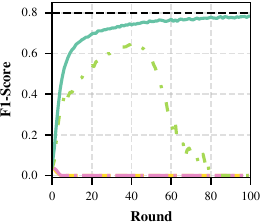}
    \caption{ALIE ($\byzFrac=0.3$)}
    \label{subfig:Purchase_CG_ALIE_f3}
  \end{subfigure}
  \end{minipage}
  \caption{F1-Score comparison between \system and three flavors of BASALT under $\byzForce=2$ and CS aggregator over the Purchase100 dataset.}~\label{fig:Purchase_CG}
\end{figure*}

Figures~\ref{fig:MNIST_CG} and~\ref{fig:Purchase_CG} illustrate the comparative performance of \system and three configurations of BASALT for Byzantine scenarios with \(\byzFrac=0.1\) and \(\byzFrac=0.3\) under FOE and ALIE attacks. The configurations of BASALT differ in the filtering threshold $\byzThreshold$. Specifically, we evaluate two settings where $\byzThreshold$ is reasonably large  compared to the expected number of Byzantine nodes in a setting with no flooding attacks (\ie, up to 6 times larger), which we describe as flexible settings. In contrast, the third configuration is an extremely conservative approach, in which
BASALT systematically filters out all models but one. Under the FOE attack (Figures~\ref{subfig:CG_FOE_f1} and~\ref{subfig:Purchase_CG_FOE_f1}), we observe that \system converges toward the optimal performance. In contrast, BASALT fails to converge for both $\byzThreshold = 4 * \byzFrac * \viewSize$ and $\byzThreshold = 6 * \byzFrac * \viewSize$ on MNIST while it experiences major fluctuations on Purchase100. The main reason behind this is that, even with large thresholds, BASALT is vulnerable to temporary spikes in $\byzFrac_{in}$. These spikes cause the number of Byzantine nodes to be higher than 6 times what one expects without flooding. While the model recovers from the attack periodically on Purchase100 (Figure~\ref{subfig:Purchase_CG_FOE_f1}), the convolutional model used on MNIST never recovers indicating extreme divergence (\eg, gradient explosion). As for the conservative version of BASALT, it achieves comparable performance to \system in $\byzFrac = 0.1$ scenarios (\eg, 91\% F1-Score versus 93\% on MNIST). Intuitively, this result is not so surprising as the conservative approach systematically filters-out more models than \system. However, this approach does not necessarily generalize as indicated by the results in the scenario where $\byzFrac = 0.3$ (Figures~\ref{subfig:CG_FOE_f3} and~\ref{subfig:Purchase_CG_FOE_f3}) where we observe that BASALT hits a brick wall as early as rounds 18 and 4, on MNIST and Purchase100, respectively, as it then rapidly deteriorates to completely diverge in the space of 30 and 16 rounds. This highlights BASALT's difficulty in scaling its resilience with higher values of $\byzFrac$.
To further investigate this, we turn our attention to the ALIE results. Figures~\ref{subfig:CG_ALIE_f1} and~\ref{subfig:Purchase_CG_ALIE_f1} show that for $\byzForce=0.1$, \system along with all flavors of BASALT converge towards the optimum at a similar rate. This contrasts with the results obtained with FOE and is ultimately down to ALIE being a stealthier attack that incorporates smaller in norm model poisons. However, with a larger number of Byzantine nodes (Figures~\ref{subfig:CG_ALIE_f3} and~\ref{subfig:Purchase_CG_ALIE_f3}), even those smaller perturbations can still cause model divergence. Particularly, we note that the flexible configurations of BASALT both diverge as soon as the 10th round in this context. Moreover, we observe a more atypical behavior for the conservative BASALT: (i) It exhibits a flat learning curve in the early up to 40 and 60 rounds, which is a direct consequence of the conservative approach consisting in aggregating only one model per round. This indicates that aggregating less models leads, as expected, to sub-optimal convergence. (ii) After that, and due to the absence of history, the seed refresh inherent to both BASALT and \system, yields a temporary increase in the proportion of Byzantine presence, leading again to Byzantine over-representation and causing the model to diverge, downgrading in F1-score from 67\% of F1-score to 0.1\% in the space of 16 rounds on MNIST. 
All in all, the takeaways from these results are threefold: First, conservatively aggregating models leads to sub-optimal performance. Second, BASALT experiences high disparities in the Byzantine presence in local views and does not scale to higher $\byzFrac$ values with the same view size. Finally, \system does not suffer from these drawbacks as it achieves near-optimal convergence in all settings. 
\subsection{Communication/Connectivity Requirements~(RQ2)}~\label{subsec:communication}
As discussed in the previous section, BASALT's robustness does not scale gracefully with the global Byzantine fraction $\byzFrac$. This limitation is in fact not inherent to BASALT but rather a consequence of the conditions in which the underlying aggregators are provably robust. Indeed, to account for bad graphs, \ie, graphs where adversaries are concentrated in sub-regions, these methods rely on having large degrees relative to the number of Byzantine nodes. Table~\ref{tab:robust_agg_comparison} illustrates these requirements in a scenario where Byzantine nodes can all be present in one neighborhood and compares them with the empirical requirements of \system. We note that $\byzThreshold$ represents an actual lower bound on the required $\viewSize$ . We use this latter to estimate the (minimal) number of messages exchanged and compute a communication cost multiplicative gain. As expected, the connectivity requirements increase at least linearly with $\byzFrac$. For example, for  $\byzFrac = 0.3$, CS demands over 180 neighbors in a 300-node graph, compared to only 60 for $\byzFrac = 0.1$. In this case, the number of messages exchanged becomes up to 9× higher than with \system. In contrast, \system achieves empirical convergence with a fixed neighborhood size ($\viewSize=20$) for both CS and GTS. This is because \system rapidly converges to a state where Byzantine nodes are uniformly dispersed across local views (mimicking $\byzFrac_{in} \approx \byzFrac$). This explains its ability to converge with communication-efficient graphs.

\begin{table}[!htp]
\centering
\caption{Comparison of the connectivity requirements for convergence under CS and GTS for a regular graph of size $\nbusers = 300$ and the empirical connectivity of \system.}~\label{tab:robust_agg_comparison}
\begin{threeparttable}
\begin{tabular}{|c|c|c|c|c|}
\hline
\multirow{2}{*}{\textbf{\begin{tabular}[c]{@{}c@{}}Robust\\Aggregator\end{tabular}}} & \multirow{2}{*}{\textbf{$\byzFrac$}} & \multirow{2}{*}{\textbf{\begin{tabular}[c]{@{}c@{}}Theoretical filtering\\ threshold \mbox{b\tnote{1}}\end{tabular}}} & \multirow{2}{*}{\textbf{\begin{tabular}[c]{@{}c@{}}Empirical filtering\\threshold (\system)\end{tabular}}} & \multirow{2}{*}{\textbf{\begin{tabular}[c]{@{}c@{}}Communication cost\\multiplicative \mbox{gain\tnote{2}}\end{tabular}}} \\
& & & & \\
\hline
\multirow{2}{*}{CS} & 0.1 & 60 & \multirow{4}{*}{$ b(t) \le 19$} & 3.1 \\
\cline{2-3} \cline{5-5}
& 0.3 & 180 & & 9 \\
\cline{1-3} \cline{5-5}
\multirow{2}{*}{GTS} & 0.1 & 30 & & 1.6 \\
\cline{2-3} \cline{5-5}
 & 0.3 & 90 & & 4.6 \\
\hline
\end{tabular}
 \begin{tablenotes}
      \footnotesize
      \item[1] Worst-case $f_{\text{in}}$; threshold required for convergence without assumptions on local Byzantine ratio.
      \item[2] Ratio between the minimal number of exchanged messages required by the theory~\cite{gaucher2024achieving} and the number of messages in \system. 
    \end{tablenotes}
\end{threeparttable}
\end{table}

\begin{figure*}[!tb]
    \centering
    \includegraphics[width=0.6\linewidth]{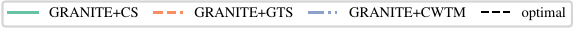}
\begin{minipage}{\textwidth}
\centering
    \begin{subfigure}[b]{0.23\linewidth}
        \includegraphics[width=\textwidth]{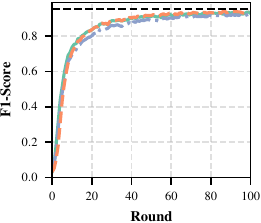}
        \caption{FOE, $\byzFrac = 0.1$}
    \end{subfigure}
    \begin{subfigure}[b]{0.23\linewidth}
        \includegraphics[width=\textwidth]{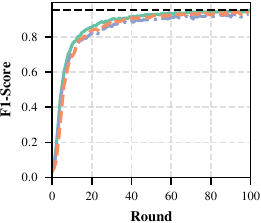}
        \caption{ALIE, $\byzFrac = 0.1$}
    \end{subfigure}
    \begin{subfigure}[b]{0.23\linewidth}
        \includegraphics[width=\textwidth]{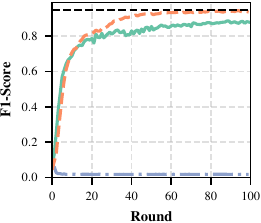}
        \caption{FOE, $\byzFrac = 0.3$}
    \end{subfigure}
    \begin{subfigure}[b]{0.23\linewidth}
        \includegraphics[width=\textwidth]{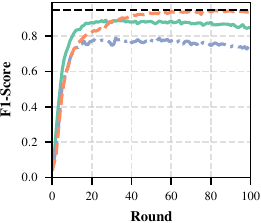}
        \caption{ALIE, $\byzFrac = 0.3$}
    \end{subfigure}
\end{minipage}
    \caption{F1-Score analysis of \system under FOE and ALIE attacks with different robust aggregators on the MNIST dataset.}
    \label{fig:aggregation_mnist}
    
\end{figure*}

\begin{figure*}[!tb]
    \centering
    \includegraphics[width=0.6\linewidth]{Plots/Aggregation_f_03_legend.pdf}
    \begin{minipage}{\textwidth}
    \centering
    \begin{subfigure}[b]{0.23\linewidth}
        \includegraphics[width=\textwidth]{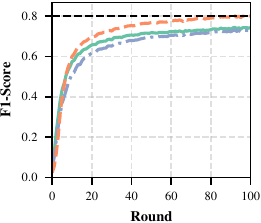}
        \caption{FOE, $\byzFrac = 0.1$}
    \end{subfigure}
    \begin{subfigure}[b]{0.23\linewidth}
        \includegraphics[width=\textwidth]{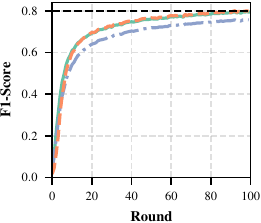}
        \caption{ALIE, $\byzFrac = 0.1$}
    \end{subfigure}
    \begin{subfigure}[b]{0.23\linewidth}
        \includegraphics[width=\textwidth]{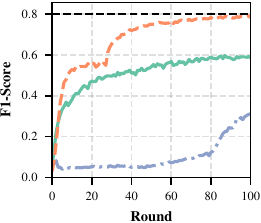}
        \caption{FOE, $\byzFrac = 0.3$\label{fig:aggregation_purchase:FOE03}}
    \end{subfigure}
    \begin{subfigure}[b]{0.23\linewidth}
        \includegraphics[width=\textwidth]{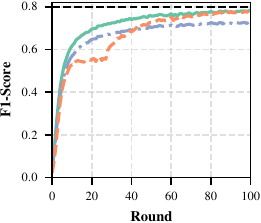}
        \caption{ALIE, $\byzFrac = 0.3$}
    \end{subfigure}
    \end{minipage}
    \caption{F1-Score analysis of \system under FOE and ALIE attacks with different robust aggregators on the Purchase100 dataset.}
    \label{fig:aggregation_purchase}
\end{figure*}

\subsection{Performance Comparison of Different Robust Aggregators~(RQ3)}
Figures~\ref{fig:aggregation_mnist} and~\ref{fig:aggregation_purchase} illustrate the F1-Score performance of \system with CS, GTS and CWTM over MNIST and Purchase100 under two Byzantine fractions. Overall, we observe that the strengths of \system are not specific to CS. For instance, they generalize to GTS, achieving near-optimal performance for both aggregators in most cases. However, in certain scenarios (most notably in Fig.~\ref{fig:aggregation_purchase:FOE03}) only GTS manages to maintain strong performance. This is likely due to CS requiring larger neighborhoods to effectively filter out malicious models, rather than any shortcoming of \system, which successfully bounded the number of Byzantine nodes in honest nodes’ views. Finally, as anticipated, CWTM proves inadequate for defending against state-of-the-art attacks, particularly in heterogeneous and fully decentralized environments.

\subsection{Effectiveness of HaPS and Quality of the Approximation \texorpdfstring{$B(t)$}{B(t)}\texorpdfstring{~(RQ4)}{ (RQ4)}}

\begin{figure}
\centering
\begin{minipage}[t]{0.48\linewidth}
    \centering
    \hspace{2.2em}\includegraphics[width=0.3\linewidth]{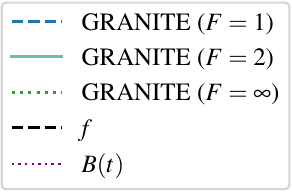}
    \includegraphics[width=0.48\linewidth]{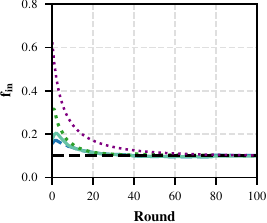}
    \caption{Evolution of $\byzFrac_{in}$ over rounds and flooding attack with different values of $\byzForce$.}\label{fig:bt}
\end{minipage}
\begin{minipage}[t]{0.48\linewidth}
\hspace{2.7em}\includegraphics[width=0.35\linewidth]{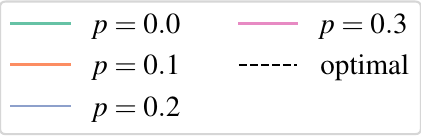}
\includegraphics[width=0.48\linewidth]{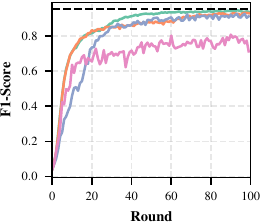}
\caption{F1-Score analysis of \system under FOE attack and four degrees of churn $p$. (\byzFrac=0.3)}\label{fig:churn}
\end{minipage}
\end{figure}

Figure~\ref{fig:bt} illustrates the evolution of $\byzFrac_{in}$ over rounds for different attack forces $\byzForce$. We observe that HaPS effectively reduces the proportion of Byzantine nodes $\byzFrac_{in}$ to a neighborhood of $\byzFrac$ within approximately 20 rounds. Moreover, $B(t)$ follows the expected  exponential decay. However, there is a large initial gap between $B(t)$ and $\byzFrac_{i}$. For instance, for $\byzForce=\infty$, where Byzantine nodes flood all honest nodes in their current view, $B(t)$ is almost 2 times larger than $\byzFrac_{i}$ (62\% versus 32\%). This gap stems from our worst-case derivation of $B(t)$, which is by design not tight in early rounds in order to remain valid even under adversaries which have unlimited communication resources or are pre-positioned in the history of honest nodes.

\subsection{Resilience to Churn~(RQ5)}
\system can be configured to tolerate churn by accounting for a maximum fraction $p$ of nodes that may leave the network. Given such a bound, nodes can conservatively compute the quantities $C(t)$, $\sigma$, and $B(t)$ under the worst-case assumption that $p \cdot \nbusers$ nodes have already departed. Figure~\ref{fig:churn} illustrates this behavior for four values of churn. We observe that \system maintains convergence to the optimum for up to $20\%$ churn. At $30\%$, the F1-score decreases from $92\%$ to $79\%$, which is primarily due to the number of honest nodes approaching the number of Byzantine nodes, rather than a limitation of \system itself.

\section{Discussion}~\label{sec:discussion}
\textbf{Resilience to Sybil Attacks.} This problem has been extensively studied, with resource-based identity mechanisms (\eg, proof of work) established as the major solution~\cite{platt2023sybil}. In contrast, most works on Byzantine-resilient Peer Sampling consider Sybil resilience to be orthogonal~\cite{auvolat2023basalt,antonov2023securecyclon,bortnikov2008brahms}. Our work adopts this common threat model. Nevertheless, there are of-the-shelf solutions that are complementary to \system. For instance, ~\cite{pilet2020foiling} propose to address nodes in a hierarchical fashion and store the views in a buffered probabilistic binary tree, which ensures that denser regions of the address space (Sybil nodes) do not have more probability of being sampled than sparse regions. Incorporating this solution in \system merely requires adopting a hierarchical addressing and a maintaining a tree-based history rather than a set. \\

\noindent\textbf{Overhead considerations.} 
Compared to existing BRPS protocols, \system is based on storing node identifiers indefinitely, which introduces an additional memory overhead. However, in a machine learning application, this cost is negligible compared to the size of a learning model. For instance, the convolutional model used in the experimental setting, modest by current standards, consists of 176k parameters. This is the equivalent of 40k IPv6 addresses. More modern models such as ResNet-18 exceeds 11M parameters, corresponding to 2.5M IPv6 addresses. The gap is even larger on the computational side where the additional operations required by \system are negligible compared to the cost of differentiation and back-propagation. Finally, we note that on other criteria such as bandwidth, \system does not incur any additional overhead compared to a traditional RPS.

\section{Related Work}~\label{sec:related_work}
In this section, we provide an overview of relevant related works in Dynamic GL, robust learning and Random Peer Sampling Protocols.
\subsection{Dynamic GL}
Recent years have witnessed both theoretical and empirical efforts to understand the advantages of Dynamic GL. Ying et al.~\cite{ying2021exponential} proved that exponential graphs, \ie, graphs with logarithmic degree and small diameter, can achieve exact-averaging in dynamic settings, improving scalability and model performance compared to static settings. In~\cite{song2022communication}, the authors show that carefully chosen dynamic topologies can yield consensus rates (\ie, the speed at which nodes agree on a shared model or value) independent of the graph size, thereby promoting sparser and communication efficient topologies. Another work~\cite{belal2022pepper} illustrates how the RPS can be tailored to build personalized neighborhoods, improving user-centric performance. More recently, several works~\cite{belal2025inferring,touat2024scrutinizing} investigated the resistance of Dynamic GL to privacy attacks and observed that the reduced attack surface of Dynamic GL coupled with its faster convergence speed enable it to gain a certain level of privacy protection. Our work build upon this literature while focusing on the robustness aspect, which has not been investigated in Dynamic GL. 
\subsection{Robust Decentralized Learning}
The literature on learning in the presence of malicious users is extensive~\cite{guerraoui2023byzantine}. Most works build on the notion of robust aggregation~\cite{blanchard2017byzantine,yin2018byzantine,yang2019byrdie,fang2022bridge}. Recently, robust aggregation evolved from comparing received models with each other to comparing each against the local model~\cite{he2022byzantine,fang2024byzantine}, showing greater effectiveness in decentralized settings. In this direction, ~\cite{gaucher2024achieving} propose two robust aggregators, proven optimal for GL over sparse topologies. Nevertheless, their guarantees 
are specific to static topologies and require optimistic initial graphs. Adopting them directly in a dynamic setting while preserving their guarantees implies much denser graphs. In this paper, we circumvent this by forcing the proportion of Byzantine nodes in local views to converge towards the global Byzantine proportion. All in all and to the best of our knowledge, our work is the first to extend Byzantine resilience to Dynamic GL in a non i.i.d. setting while also considering adversaries that target the Random Peer Sampling protocol.

\subsection{Byzantine-resilient Random Peer Sampling}
Several Byzantine-resilient RPS protocols have emerged in the literature~\cite{auvolat2023basalt,jesi2010secure,bortnikov2008brahms}.  \cite{jesi2010secure} proposed a protocol where nodes collaboratively collect robust statistics about the network in order to detect misbehaving nodes. This approach is dedicated to Hub Attacks and suffers from a cold start problem. In Brahms~\cite{bortnikov2008brahms}, the authors propose to include a sample of previously encountered nodes in the computation of views. This allows Brahms ensure probabilistic resilience against Byzantine influence. Recently, BASALT~\cite{auvolat2023basalt} extended the threat model of Brahms by considering adversaries with unlimited communication power. With careful parameterization, BASALT has been shown to yield an exponentially decaying probability of node isolation. In this work, we consider BASALT as a competitor, and show that an exponentially decaying probability of isolation is not sufficient to achieve stable and satisfactory convergence.

\section{Conclusion}
\label{sec:conclusion_granit}

Byzantine resilient decentralized learning is an appealing alternative to classical Federated Learning from various perspectives (\eg, scalability, decentralized governance, resilience). However, existing solutions often assume that Byzantine nodes do no cheat in the underlying networking layer (\ie, in the peer sampling protocol).
In this work, we are the first to tackle a critical blind spot at the intersection of decentralized learning and secure peer sampling: the incompatibility between existing Byzantine-resilient Peer Sampling (BRPS) protocols and the requirements of robust Gossip Learning. We showed that even though robust aggregators can mitigate poisoning in ideal conditions, they are vulnerable when Byzantine nodes manipulate the network layer to be over-represented in the view of honest nodes, leading to convergence failures even under moderate adversarial presence.

To address this challenge, we introduced \system, a novel framework that bridges robust aggregation with secure, history-aware peer sampling (HaPS). \system also uses APT that dynamically adjusts aggregation filters based on time-evolving adversarial presence. We provided a formal analysis of the exponential decay of Byzantine proportions in local views and leveraged this to guarantee (with high probability) the effectiveness of robust aggregators.

Empirical evaluations against the strongest state-of-the-art attacks demonstrated that \system not only preserves the communication efficiency and scalability of Dynamic GL but also significantly enhances its resilience, enabling convergence with up to 30\% Byzantine nodes and operating effectively on sparse topologies.

\bibliographystyle{plain}
\bibliography{biblio}

\appendix

\section{Proof of Theorem~\ref{thm:convergence-c}}
\label{appendix:ode-derivation}
Given our assumptions, we have:
\begin{equation}
    B(t) = \frac{\byz}{\byz + \correctp{t}}, \qquad 
    B(0) = \frac{\byz}{\byz + \correctp{0}}
\end{equation}
The growth in $\correctp{t}$ is driven by the rate $\sigma$ at which (possibly repeated)
honest identifiers are added to a history through push and pull interactions:
\begin{equation}
    \sigma = \sigma_{\text{pull}} + \sigma_{\text{push}}.
\end{equation}

\textbf{Pull.}  
A node pulls from a neighbor selected uniformly at random.  
The probability that this neighbor is honest is at least $\frac{\correctp{0}}{\correctp{0} + \byz}$.
Moreover, an honest node's own view contains, on average, the same honest fraction.
Since each view has size $\viewSize$, the expected number of correct identifiers
received from a pull is at least:
\begin{equation}
    \sigma_{\text{pull}} 
    = \left(\frac{\correctp{0}}{\correctp{0} + \byz}\right)^2 \cdot \viewSize.
\end{equation}

\textbf{Push.}  
Each node receives push messages from other nodes in the network.
A node expects to receive $\frac{|\honest|}{\nbusers - 1}$
push requests from honest nodes, each contributing  
$\viewSize \cdot \frac{\correctp{0}}{\correctp{0} + \byz}$  
correct identifiers on average. Thus:
\begin{equation}
    \sigma_{\text{push}}
    = \frac{|\honest|}{\nbusers - 1}
    \cdot \frac{\correctp{0}}{\correctp{0} + \byz} \cdot \viewSize.
\end{equation}
Note that both $\sigma_{\text{pull}}$ and $\sigma_{\text{push}}$ are evaluated 
at $t=0$, providing a conservative lower bound on the true arrival rate, 
which increases as honest identifiers accumulate in histories over time.

Each incoming identifier is unique with probability  
$\frac{|\honest| - \correctp{t}}{|\honest|}$,  
leading to the following ODE:
\begin{equation}
    \frac{d\correctp{t}}{dt}
    = \sigma \cdot \frac{|\honest| - \correctp{t}}{|\honest|}.
    \label{eq:ode}
\end{equation}
Rearranging~\eqref{eq:ode} into standard linear form:
\begin{equation}
    \frac{d\correctp{t}}{dt} + \frac{\sigma}{|\honest|}\,\correctp{t} = \sigma.
\end{equation}

For some constant \(K\), the general solution is:
\begin{equation}
    \correctp{t} = |\honest| + K\exp\!\left(-\frac{\sigma}{|\honest|}t\right).
\end{equation}

Applying the initial condition \(\correctp{0}\) gives:
\begin{equation}
    \correctp{t} = |\honest| - \left(|\honest|-\correctp{0}\right)\exp\!\left(-\frac{\sigma}{|\honest|}t\right).
\end{equation}

Substituting into $B(t)$ we have:
\begin{equation}
    B(t) = \frac{\byz}{\byz + |\honest| - \left(|\honest|-\correctp{0}\right)\exp\!\left(-\frac{\sigma}{|\honest|}t\right)}.
\end{equation}
    
This completes the derivation of Theorem~\ref{thm:convergence-c}.

\section{Proof of Theorem~\ref{thm:robustness_threshold}}
\label{app:derivation}
At round $t$, let $\graphhpu{t}$ denote the honest subgraph, with adjacency matrix $A_t$ and degree matrix $D_t$, and let $W_t = D_t - A_t$ be its (unnormalized) Laplacian. Let $\viewSize$ denote the expected honest degree scale, and let $B(t)$ be the expected Byzantine proportion. We recall the following classical result (see theorem 1.1 in \cite{feige2005spectral}):

\begin{theorem}[Spectral bound for random graphs]
\label{thm:er_spectral_appendix}
Let $G \sim \mathcal{G}(n,p)$ with
\[
c_0 \frac{\log n}{n} \le p \le \frac{n^{1/3}}{n (\log n)^{5/3}}.
\]
Then, for every $c>0$, there exists a constant $c'>0$ such that, with probability at least $1 - n^{-c}$,
\[
\lambda_2(A(G)) \le c'\sqrt{np},
\]
where $\lambda_2(A(G))$ denotes the largest non-trivial eigenvalue of the adjacency matrix.
\end{theorem}

By definition of the Laplacian,
\[
W_t = D_t - A_t,
\]
and the algebraic connectivity satisfies
\begin{align}
\lambda_2(W_t) &= \deg_{\mathrm{avg}}(\graphhpu{t}) - \lambda_2(A_t) \notag\\
&= \viewSize (1-B(t)) - \lambda_2(A_t).
\end{align}

Applying Theorem~\ref{thm:er_spectral_appendix} gives, with high probability,
\begin{align}
\lambda_2(W_t) &\ge \viewSize (1-B(t)) - c'\sqrt{n p_t} \notag\\
&= \viewSize (1-B(t)) - c'\sqrt{n (1-B(t))}.
\label{eq:appendix_laplacian_bound}
\end{align}

The robustness requirement of \system imposes
\[
\lambda_2(\graphhpu{t}) \ge 2 b(t), \qquad b(t) = \viewSize (1+\delta) B(t),
\]
where $\delta \ge 0$ is the confidence parameter. Combining this with \eqref{eq:appendix_laplacian_bound}, we obtain
\begin{align}
\viewSize (1-B(t)) - c'\sqrt{n (1-B(t))} &\ge 2 b(t) \notag\\
\viewSize (1-B(t)) - c'\sqrt{n (1-B(t))} &\ge 2 \viewSize (1+\delta) B(t) \notag\\
B(t) &\le \frac{1 - \frac{c'}{\viewSize} \sqrt{n(1-B(t))}}{3 + 2\delta}.
\end{align}

Neglecting the lower-order spectral term $c'\sqrt{n(1-B(t))}/\viewSize$ recovers the asymptotic threshold
\begin{equation}
B(t) < \frac{1}{3+2\delta}.
\end{equation}

This completes the proof.
\section{Proof of Lemma~\ref{lem:robustness_convergence}}
\label{app:robustness_convergence}
We have shown that $B(t)$ evolves according to
\begin{equation}
B(t) = \frac{B}{B + n_h - (n_h - \correctp{0}) \exp\left(-\frac{\sigma}{n_h} t\right)}.
\end{equation}

Robustness is guaranteed once $B(t) \leq B_{\mathrm{target}}$, where $B_{\mathrm{target}}$ denotes the threshold established in Theorem~\ref{thm:robustness_threshold}. Since $B(t)$ is monotonically decreasing and continuous, such a finite $T$ always exists. To characterize its scaling, observe that $B(t)$ is driven by the exponential term $\exp(-\sigma t / n_h)$, which decays on the timescale $t \sim n_h / \sigma$. Taking logarithms of the exact condition $B(T) \leq B_{\mathrm{target}}$ yields:
\begin{equation}
T = \frac{n_h}{\sigma} \ln \frac{n_h - \correctp{0}}{n_h + B - \frac{B}{B_{\mathrm{target}}}}.
\end{equation}

Under typical system parameters where $B \ll n_h$ and $\correctp{0} \ll n_h$, the argument of the logarithm is $O(1)$ up to constants, giving
\begin{equation}
T = O\!\left(\frac{n_h}{\sigma}\right).
\end{equation}

which completes the proof.

\end{document}